\documentclass[runningheads]{llncs}
\usepackage[pdftex]{graphicx}
\usepackage{amsmath,amssymb} 
\usepackage{color}
\usepackage[width=122mm,left=12mm,paperwidth=146mm,height=193mm,top=12mm,paperheight=217mm]{geometry}

\usepackage{booktabs}
\usepackage{bm}

\newcommand{\Tref}[1]{Table~\ref{#1}}

\newcommand{\Fref}[1]{Fig.~\ref{#1}}

\begin{document}
\pagestyle{headings}
\mainmatter
\def\ECCV18SubNumber{1654}  

\title{Unsupervised Adversarial Learning of 3D Human Pose from 2D Joint Locations} 

\titlerunning{Unsupervised Adversarial Learning of 3D Human Pose}

\authorrunning{Kudo et al.}

\author{
Yasunori Kudo\inst{1}, Keisuke Ogaki\inst{2}, Yusuke Matsui\inst{3}, and Yuri Odagiri\inst{2}
}
\institute{
{Keio University, Japan}\and
{DWANGO Co., Ltd., Japan}\and
{National Institute of Informatics, Japan}
}

\maketitle

\begin{abstract}
The task of three-dimensional (3D) human pose estimation from a single image can be divided into two parts: (1) Two-dimensional (2D) human joint detection from the image and (2) estimating a 3D pose from the 2D joints.
Herein, we focus on the second part, i.e., a 3D pose estimation from 2D joint locations.
The problem with existing methods is that they require either (1) a 3D pose dataset or (2) 2D joint locations in consecutive frames taken from a video sequence.
We aim to solve these problems.
For the first time, we propose a method that learns a 3D human pose without any 3D datasets.
Our method can predict a 3D pose from 2D joint locations in a single image.
Our system is based on the generative adversarial networks, and the networks are trained in an unsupervised manner.
Our primary idea is that, if the network can predict a 3D human pose correctly, the 3D pose that is projected onto a 2D plane should not collapse even if it is rotated perpendicularly.
We evaluated the performance of our method using Human3.6M and the MPII dataset and showed that our network can predict a 3D pose well even if the 3D dataset is not available during training.
\end{abstract}

\section{Introduction}
Given a single image, a three-dimensional (3D) human pose estimation, defined as a problem in the localization of human joints in a 3D space, is an important and challenging problem in computer vision.
It can be applied to many areas, including action recognition, virtual reality, and human--computer interaction.
A common approach divides this problem into two parts.
First, two-dimensional (2D) human joint locations are estimated from an image.
Next, a 3D pose is predicted from the 2D joint locations.
This approach is known as the ``pipeline approach.''
The first part can be achieved by a joint detector such as in \cite{toshev2014deeppose,tompson2014joint,chen2014articulated,tompson2015efficient,fan2015combining,wei2016cpm,carreira2016iterative,bulat2016heatmap,pishchulin2016deepcut,newell2016stacked,cao2017realtime}.
In our research, we focus on the second part, i.e., a 3D pose estimation from 2D joint locations.
This task is particularly challenging because reconstructing 3D information from 2D joint locations is essentially an ill-posed problem.

Two approaches can be used to estimate a human 3D pose from 2D joint locations.
The first approach uses a 3D dataset such as \cite{sigal2010humaneva,ionescu2014human3.6m,mpi-inf}.
These datasets have human joint coordinates in a specific 3D space, which is captured by the mocap system.
Typical methods~\cite{martinez20173dbaseline,tome2017lifting,wu2016single,pavlakos2017coarse} in this approach train neural networks to estimate a 3D pose from 2D joint locations.
The second approach uses only 2D human joint locations to predict a 3D pose.
For example, the non-rigid structure from motion (NRSfM) approach~\cite{bregler2000recovering,akhter2011trajectory,gotardo2011computing,lee2016procrustean} uses known 2D correspondences for the joints in multiple images from a monocular video or multiview cameras to recover a 3D pose.

However, some problems exist on these approaches.
The first 3D approach requires a large amount of 3D data for training.
Moreover, if we estimate a 3D pose via a pipeline approach that uses this 3D approach, the joints of the resulting 3D pose depend totally on the joints provided by the 3D datasets.
In other words, the pipeline approach cannot estimate the 3D locations of joints that are detected by a 2D joint detector.
For example, if a 2D joint detector can detect ears but a 3D dataset does not have an annotation of ears, the pipeline approach cannot estimate the 3D locations of ears.
The second 2D approach such as the NRSfM does not require any 3D datasets.
However, it requires a large number of 2D joint locations from many images at test time (e.g., consecutive frames taken from a video sequence) to recover a 3D pose correctly.

\begin{figure}[tb]
  \begin{center}
    \includegraphics[width=12cm]{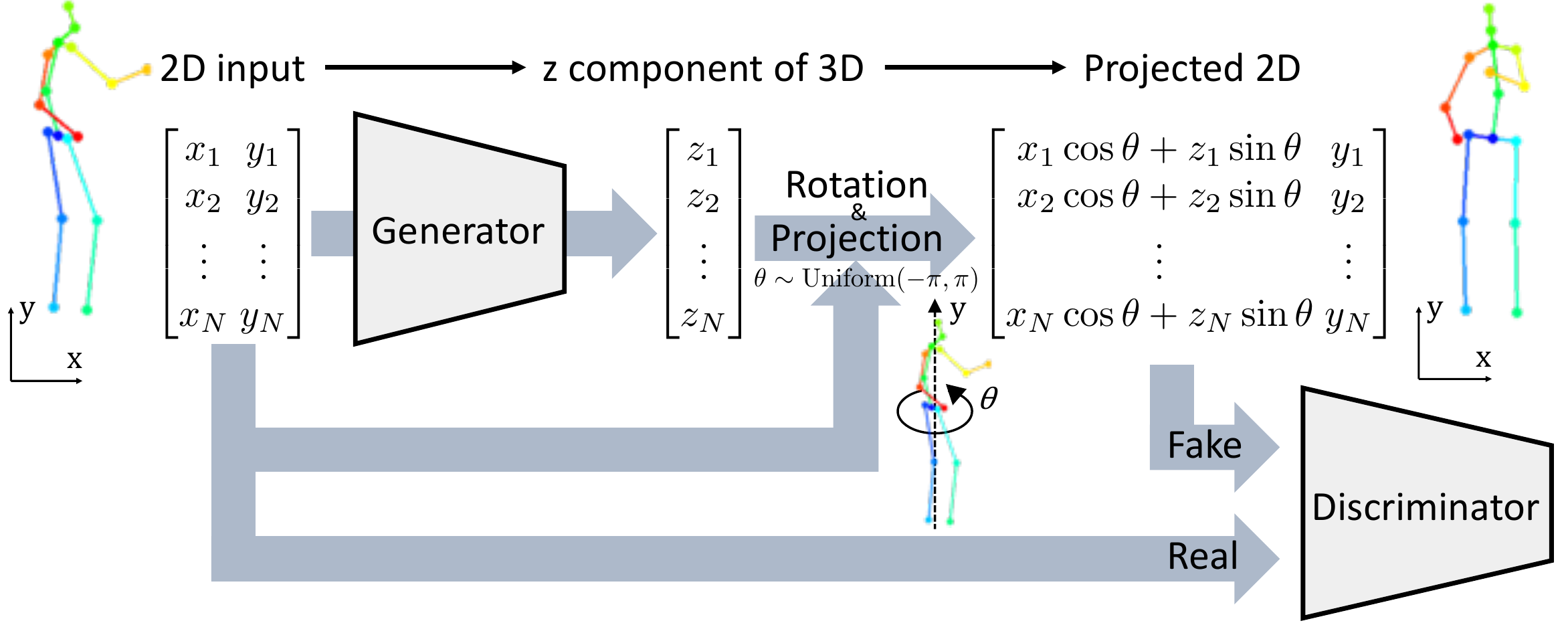}
    \caption{
Framework of our approach.
Given the 2D joint locations in a single image, the generator estimates the $z$-component for each input joint.
The generator estimates the $z$-component of a 3D pose given the single 2D joint locations. 
The resulting 3D pose is rotated around the $y$-axis by $\theta$ radian, which is obtained from a uniform distribution on $[-\pi, \pi]$.
The rotated pose is subsequently projected onto the $xy$-plane.
The discriminator is trained to distinguish between a real 2D pose and a projected 3D pose.
In this approach of training a 3D pose from a 2D pose, 3D datasets were not used.
}
    \label{system}
  \end{center}
\end{figure}

Herein, we propose a simple yet novel approach to predict a 3D human pose from 2D joint locations in a single image using a neural network.
The network is trained in an unsupervised manner without any 3D signals. 
\Fref{system} shows the overview of our approach.
Our approach is based on the generative adversarial networks~\cite{goodfellow2014generative} that predict a 3D pose ($xyz$ coordinates) given 2D human joint locations.
Our primary idea is that, if the network can predict joints correctly,
\textit{an output 3D pose should not collapse even if it is rotated perpendicularly}.
Let the $x$-axis be the vertical axis and the $y$-axis be the horizontal axis in an image.
Given the 2D joint locations on the $xy$-plane, the generator is trained to estimate the $z$-component for each joint.
The generated 3D pose is rotated around the $y$-axis by $\theta$ radian that is obtained from a uniform distribution on $[-\pi, \pi]$.
The rotated pose is then projected onto the $xy$-plane.
If the generator estimates a $z$-component for each joint correctly, all projected 3D poses are indistinguishable from real 2D data.
This is because we assume that the 2D dataset contains many 2D poses obtained from various angles around the $y$-axis.
Therefore, in our system, the discriminator is trained to distinguish between a real 2D pose and a projected 3D pose.
In a test phase, new input 2D joints are fed into the generator, and the corresponding $z$-components are predicted.

We tested our method using the Human3.6M dataset, which is widely used as a benchmark for 3D human pose estimation from 2D joint locations.
This test reveals that if many ground truth 2D joint locations or detected 2D joint locations are available, we can estimate the 3D human pose accurately.
We applied our method to the MPII dataset as well.
This dataset does not have 3D annotations, but 2D annotations only, because the images in this dataset are taken in the wild, not in controlled environments like the Human3.6M.
We show that our approach works well for 2D datasets as well.

This work presents the following contributions:
\begin{itemize}
\item For the first time, we propose an unsupervised method that learns a 3D human pose from 2D joint locations in a single image without any 3D datasets.
\item Our approach is applicable to the problem of 3D pose estimation from a single image using a 2D joint detector. We can estimate the 3D locations of joints that can be detected by a 2D joint detector.
\end{itemize}

\section{Related Work}


\subsubsection{3D human pose estimation.}
If a large number of pairs (images and 3D pose) are available, the standard approach to the problem of 3D human pose estimation from a single image is by supervised learning.
Li and Chan~\cite{li20143dpose} trained neural networks to directly regress 3D joint locations from an image.
Zhou et al.~\cite{zhou2016deepkinematic} used the prior knowledge on 3D human constraints while training neural networks to guarantee the geometric validity of a 3D human pose.
Tekin et al.~\cite{tekin2016structured} trained an auto-encoder network for a 3D human pose.
Moreover, they trained another network to estimate a 3D pose from an image using the high-dimensional latent pose representations from the auto-encoder as supervision.
When training a 3D pose using these approaches, a 3D pose dataset is required.
However, the images in the existing 3D datasets~\cite{sigal2010humaneva,ionescu2014human3.6m,mpi-inf} are captured in controlled environments with the mocap system.
The models trained on these datasets cannot be generalized well to other environments, such as those in the wild.

\subsubsection{3D pose from 2D joint positions.}
Many researchers have attempted to train and estimate the 3D human pose from an image captured in the wild. The standard approach divides this problem into two parts. 
First, 2D joint locations are estimated from an image by a 2D joint detector~\cite{toshev2014deeppose,tompson2014joint,chen2014articulated,tompson2015efficient,fan2015combining,wei2016cpm,carreira2016iterative,bulat2016heatmap,pishchulin2016deepcut,newell2016stacked,cao2017realtime} trained using 2D datasets, i.e., the datasets that have a large number of human images and 2D joint locations.
Many large 2D pose datasets are available~\cite{mpii,mscoco,flic}, whose images are captured in the wild because they are much easier to create than those from 3D pose datasets.
The second step utilizes the estimated 2D joint locations and trains to reconstruct a 3D pose using 3D datasets such as~\cite{sigal2010humaneva,ionescu2014human3.6m}.
Martinez et al.~\cite{martinez20173dbaseline} trained neural networks simply to regress a 3D pose from a detected 2D pose in a specific 3D space.
Other approaches~\cite{tome2017lifting,wu2016single,pavlakos2017coarse} created 2D joint probability heatmaps from the detected 2D joints and trained convolutional neural networks to estimate a 3D pose from these heatmaps.
The NRSfM approaches~\cite{bregler2000recovering,akhter2011trajectory,gotardo2011computing,lee2016procrustean} recover 3D poses from 2D joint locations in a monocular video.
The advantage of this approach is that 3D datasets are not required when estimating a 3D pose from 2D joint locations; instead, a large number of human 2D poses that are obtained from multiview cameras or a sequence of a monocular videos are required.
Our approach is different from these methods of lifting 2D to 3D because we require 2D joints from a single image (not multiple images), without any 3D datasets.

\subsubsection{End-to-end approaches with 2D and 3D datasets.}
Unlike the approach that divides the problem of 3D human pose estimation into two parts, some end-to-end approaches are available for the 3D human pose estimation in the wild.
Mehta et al.~\cite{mpi-inf} pre-trained the convolutional neural networks to estimate 2D joint locations from an image using 2D datasets and fine-tuned the networks to predict a 3D pose using 3D datasets in a supervised manner.
Zhou et al.~\cite{zhou2017towards} proposed a weakly supervised transfer learning method that trains a unified convolutional neural network to estimate 2D joint locations and their depths using mixed 2D and 3D datasets. 
\cite{hmrKanazawa17,tung2017adversarial} proposed the methods that do not need the paired data of a 3D pose and its image.
They trained the generative adversarial networks~\cite{goodfellow2014generative} using paired 2D datasets and unpaired 3D poses such that the discriminator network can learn the constraints of 3D human poses.
Those approaches are similar to our work in that they train the generative adversarial networks; however, unlike those approaches, our method can learn the constraints of a 3D human pose from only 2D joint locations.
Moreover, the joints of the resulting 3D pose predicted by those approaches that use both 2D and 3D datasets are in the different positions from the joints provided by the 2D datasets.

\section{Proposed Method}
In this section, we introduce our method that creates models of human pose in the following relative 3D space ($xyz$-coordinates), given 2D joint locations.
First, a specific human joint (``central hip'' in the Human3.6M dataset~\cite{ionescu2014human3.6m} for example) locates itself at the origin point in this space.
Next, the orthographic projection of the 3D joints onto the $xy$-plane must be input to 2D joint locations.
Therefore, our goal is to estimate the $z$-component for each human joint in this 3D space.

Our system assumes that an output 3D pose should be robust against the rotation around the $y$-axis if the $z$-components are estimated correctly.
Therefore, if the output 3D pose is rotated and projected onto the 2D plane, the result is supposed to be natural.
The discriminator is trained to distinguish between a real 2D pose and a projected 3D pose.
The generator is trained to fool the discriminator.
The generator trained through this system is supposed to generate the corresponding $z$-components to input the 2D joint locations accurately.

\subsection{Generative adversarial networks for 3D pose estimation}
We aim to learn the function $G: \mathbb{R}^{N \times 2} \rightarrow \mathbb{R}^N$ that takes a 2D pose 
${\bm p} =
\begin{bmatrix}
{\bm p}_1 & \dots & {\bm p}_N
\end{bmatrix}^{\top}=
\begin{bmatrix}
x_1 & \dots & x_N \\
y_1 & \dots & y_N
\end{bmatrix}^{\top}
\in \mathbb{R}^{N \times 2}$ 
as an input and returns 
${\bm z}=
\begin{bmatrix}
z_1 & \dots & z_N
\end{bmatrix}^{\top}
\in \mathbb{R}^N$ where $N$ denotes the number of joint locations.
For example, $N$ is set as 17 for the Human3.6M dataset~\cite{ionescu2014human3.6m} and 16 for the MPII dataset~\cite{mpii}. Each joint represents the spine, head, shoulder, hip, etc.

\Fref{system} shows the overview of our method. 
The generator is defined as a function $G$, and estimates ${\bm z}=G({\bm p})$ of a 3D pose.
The generated 3D pose is rotated around the $y$-axis by $\theta$ radian that is obtained from a uniform distribution on $[-\pi, \pi]$ and projected onto $xy$-plane.
The projected 3D pose 
$\hat{\bm p} =
\begin{bmatrix}
\hat{\bm p}_1 & \dots & \hat{\bm p}_N
\end{bmatrix}^{\top} \in \mathbb{R}^{N \times 2}$
satisfies the following equation:
\begin{eqnarray}
  \hat{\bm p} = 
  f({\bm p}, {\bm z}; \theta) = 
  {\bm p}
  \begin{bmatrix}
  \cos \theta & 0 \\
  0 & 1
  \end{bmatrix}
  + {\bm z}
  \begin{bmatrix}
  \sin \theta & 0
  \end{bmatrix}
\end{eqnarray}
where $f$ denotes the rotation and the projection function. Because the manipulations of the rotation and the projection are differentiable, the gradients of the parameters of $G$ can be calculated by the back propagation algorithm.

The discriminator is defined as the function $D : \mathbb{R}^{N \times 2} \rightarrow \mathbb{R}$.
It is trained to distinguish between ${\bm p}$ (2D pose that are the inputs to the generator) and $\hat{\bm p}$ (projected 3D pose from ${\bm z}$ estimated by the generator).
We use the following standard objective function for $G$ and $D$:
\begin{eqnarray}
V(G, D) = \mathbb{E}_{\bm p}[\log D({\bm p})] + \mathbb{E}_{{\bm p}, \theta}[\log (1 - D(f({\bm p}, G({\bm p}); \theta)))].
\label{gan-objective}
\end{eqnarray}
The discriminator is optimized to maximize $V$ and the generator is optimized to minimize $V$.

\subsection{Network design}
\begin{figure}[tb]
  \begin{center}
    \includegraphics[width=12cm]{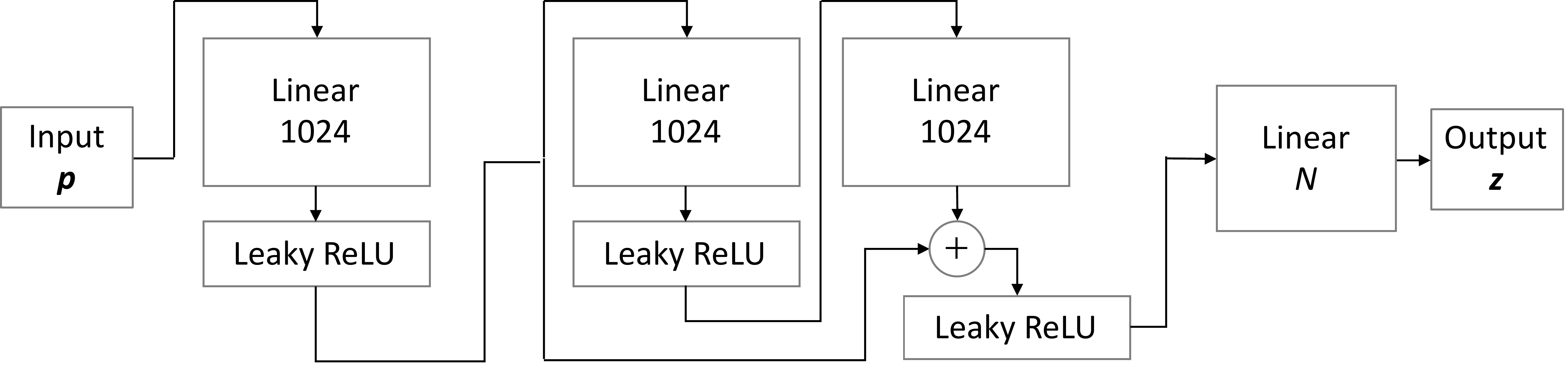}
    \caption{
Overview of the generator in our method.
This network consists of four linear layers, leaky relu functions, and skip connections.
The input to the generator is a 2D joint location ${\bm p}$ and the output is the $z$-components for each joint.
The discriminator has the same network design as the generator except that the number of units of the final layer is one.
}
    \label{gen}
  \end{center}
\end{figure}

The design of neural networks is similar to the design proposed by Martinez et al.~\cite{martinez20173dbaseline}. \Fref{gen} shows the overview of the generator in our method.
Both the generator and the discriminator consist of four linear layers and leaky relu functions~\cite{maas2013rectifier,xu2015empirical} after each layer except the final layer.
The number of units for each hidden layer is 1024.
We also introduce the skip connections~\cite{resnet}.
Although Martinez et al.~\cite{martinez20173dbaseline} used batch normalization~\cite{batchnorm}, dropout~\cite{dropout} and relu functions~\cite{relu}, we found that these regularization techniques were not effective in training our generative adversarial networks.

\subsection{Normalization of 2D joint locations}
As is widely known, the normalization of 2D joint locations is critical for the performance of 3D pose estimation~\cite{martinez20173dbaseline,tome2017lifting,zhou2017towards}.
In our case, the normalization is also important to train the generator and the discriminator efficiently because the normalization reduces the input data entropy and the divergence between the input data and projected 3D pose.
Thus, we apply the following normalization steps to all joints before the training.
First, we set a specific joint as the human central joint.
Next, we subtract the value of the central joint coordinate from all joints.
Finally, we divide the value of all joint coordinates by the scale factor, defined as the mean Euclidean distance of all joints from the central joint.
In the training phase, this relative coordinate is used to represent each position (${\bm p}$).
We used the ``central hip'' as a central joint in the Human3.6M dataset and the MPII dataset.

\subsection{Constraints of 3D human poses}
\begin{figure}[tb]
  \begin{center}
    \begin{tabular}{c}
      \begin{minipage}{0.6\hsize}
        \begin{center}
          \includegraphics[clip, width=7cm]{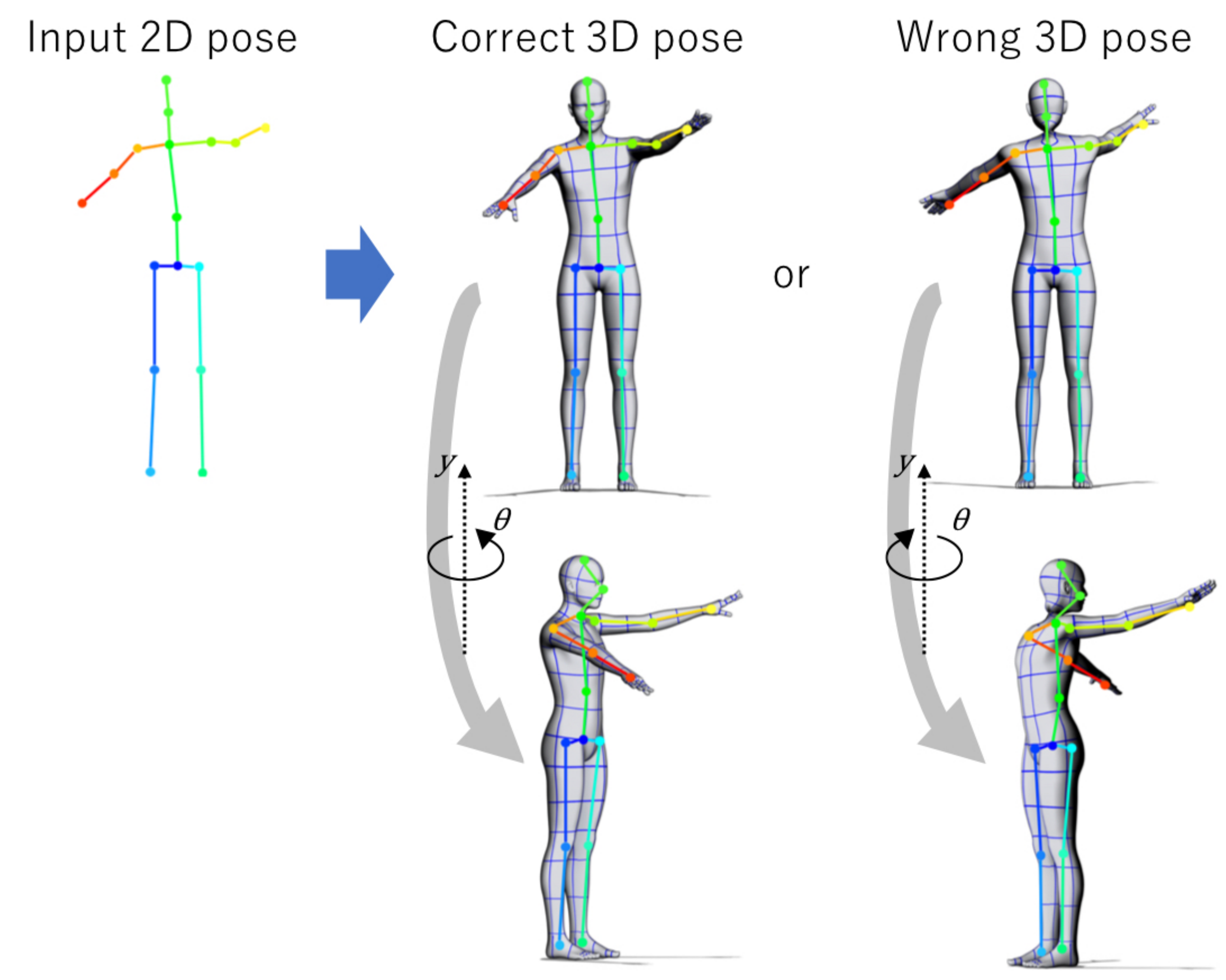}
          \caption{
          2D pose discriminator cannot detect fake 3D poses whose all $z$-components are inversion of ones of true 3D pose.
          This is because the projection of ``correct 3D pose'' that is rotated around $y$-axis by $\theta$ radian is the same 2D pose as the projection of ``wrong 3D pose'' that is rotated by $-\theta$ radian.
          The discriminator cannot distinguish these two 2D poses.
          }
    	  \label{pose}
        \end{center}
      \end{minipage}

      \begin{minipage}{0.06\hsize}
      \includegraphics[clip, width=0.1cm]{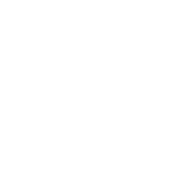}
      \end{minipage}

      \begin{minipage}{0.3\hsize}
        \begin{center}
          \includegraphics[clip, width=3.5cm]{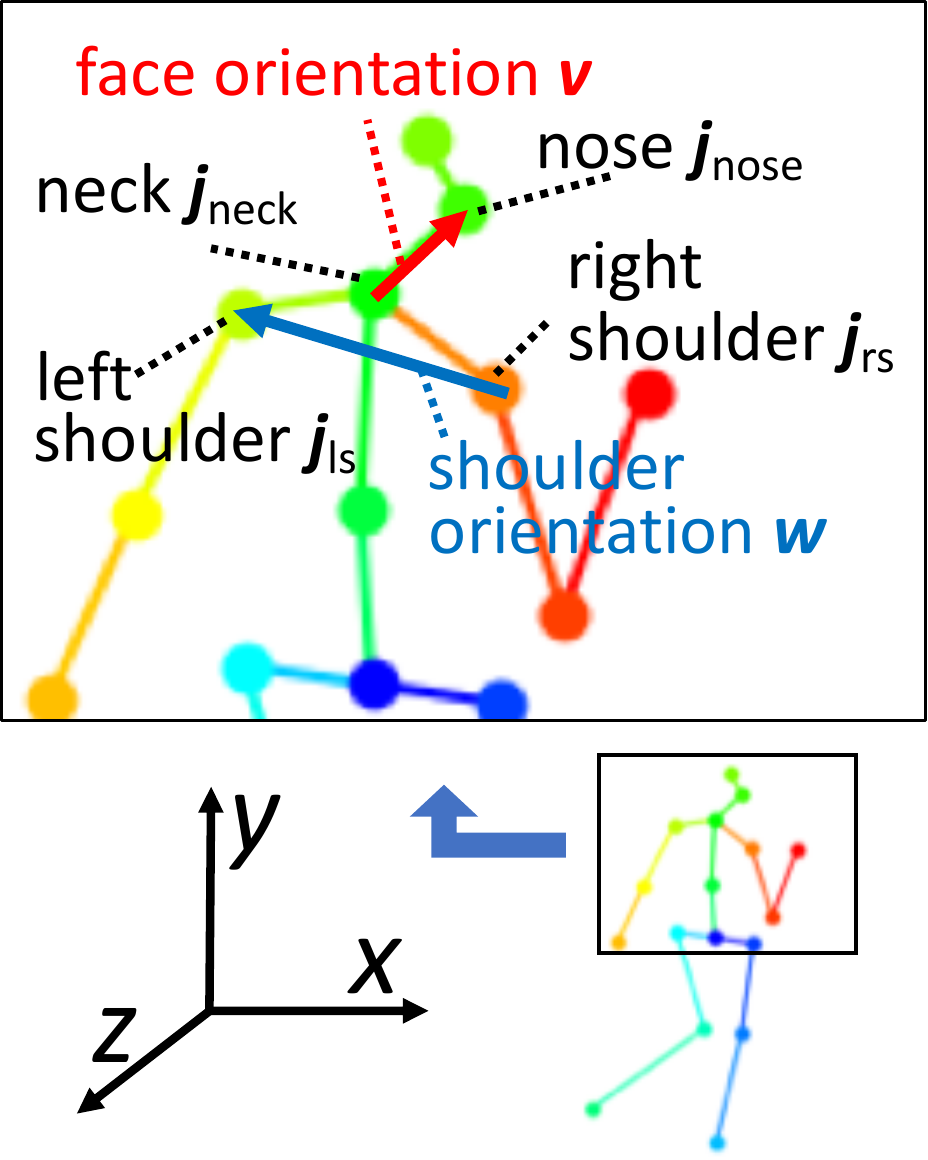}
          \caption{Heuristic loss of the constraints of human pose is used.
          The angle $\beta$ between the face orientation and the shoulder orientation on the $zx$-plane should satisfy $\sin \beta \geq 0$.
          }
    	  \label{joint}
        \end{center}
      \end{minipage}

    \end{tabular}
  \end{center}
\end{figure}

We found that our training system cannot discriminate between 3D poses and those whose $z$-components are inverted, while inverted 3D poses are impossible for human (\Fref{pose}).
After training, the generator results to generate inverted poses for about a half of inputs.
To avoid this ambiguity, we command the ``correct 3D poses'' and the ``wrong 3D poses'' to the generator by introducing a new heuristic loss function.
This loss function is based on the assumption that the ``right shoulder'' joint should be on the right side of the head.

In the Human3.6M dataset, we define the face orientation vector as ${\bm v} = [v_x, v_y, v_z] = {\bm j}_{\rm{nose}} - {\bm j}_{\rm{neck}} \in \mathbb{R}^3$ and the shoulder orientation vector as ${\bm w} = [w_x, w_y, w_z] ={\bm j}_{\rm{ls}} - {\bm j}_{\rm{rs}} \in \mathbb{R}^3$ where ${\bm j}_{\rm{nose}}, {\bm j}_{\rm{neck}}, {\bm j}_{\rm{ls}}, {\bm j}_{\rm{rs}} \in \mathbb{R}^3$ denote the 3D joint locations of the nose, neck, left shoulder, and right shoulder, respectively (\Fref{joint}).
To satisfy the constraint above, the angle $\beta$ between ${\bm v}$ and ${\bm w}$ on the $zx$-plane must satisfy $\sin \beta = \frac{v_z w_x - v_x w_z}{{\| {\bm v} \| \| {\bm w} \| }} \geq 0$ .
To satisfy this inequality, we introduce a new loss function:
\begin{eqnarray}
L_{\rm angle} = \max \left( 0, -\sin \beta \right) = \max \left( 0, \frac{v_x w_z - v_z w_x}{{\| {\bm v} \| \| {\bm w} \| }} \right).
\end{eqnarray}
The final objective function with this new angle loss is
\begin{eqnarray}
V(G, D) = \mathbb{E}_{\bm p}[\log D({\bm p})] + \mathbb{E}_{{\bm p}, \theta}[\log (1 - D(f({\bm p}, G({\bm p}); \theta)))+L_{\rm angle}].
\end{eqnarray}

\section{Experimental Evaluation}
\subsection{Datasets and protocols}
We evaluated our method using a standard dataset for the 3D human pose estimation\footnote{Further results using the MPI-INF-3DHP dataset~\cite{mpi-inf} are included in the supplemental material.}: Human3.6M~\cite{ionescu2014human3.6m}.
We also show qualitative results on the MPII dataset~\cite{mpii} that provides only 2D annotations.

The Human3.6M dataset is the largest dataset for the 3D human pose. It consists of 3.6 million images captured by four RGB cameras and the mocap system in controlled lab environments.
In this dataset, seven professional actors perform 15 actions, such as discussing, walking, taking photo, and eating.
Each actor has 32 joints; however, we used 17 joints for training and testing following previous works.
We evaluated our method by the Human3.6M standard protocol.
This protocol uses subjects 1, 5, 6, 7, and 8 for training and 9, 11 for testing.
We report the average error in millimeters between the ground truth 3D pose and our predictions across all joints and cameras in the space relative to the ``central hip'' joint.

The MPII dataset, which provides 2D annotations only, consists of 2.5 million images for training and 3000 images for testing.
The humans in the images are annotated with 16 joints.
Unlike the Human3.6M dataset, the images in the MPII dataset are captured in the wild.
Therefore, the models that were trained using such images are supposed to be generalized well.
Although the MPII provides images in the wild, it provides only 2D annotations because annotating 3D joint locations to images in the wild is challenging.
Because the images are also not always taken from a monocular video or multiview cameras, the NRSfM approach cannot be used.
We show that our approach performs well using only 2D datasets, whose images are captured from a monocular camera such as those from the MPII dataset.

\subsection{Training details}
To optimize the generator and the discriminator, we used Adam~\cite{adam} and a mini-batch size of 16.
We trained the networks for 50 epochs.
The initial weight of each layer was obtained from a Gaussian distribution with a standard deviation of 0.14.
We updated the generator and the discriminator at every iteration, but we did not update the discriminator if its accuracy was greater than 0.9.
The generator was not updated if the accuracy of the discriminator was less than 0.1.
This update rule enabled the networks to train with stability.
Our system is implemented by Chainer~\cite{chainer} and the training time for 50 epochs was approximately 8 hours using the Intel Core i7-5960X CPU, Nvidia Titan X GPU, 32-GB RAM.

\subsection{3D pose from ground truth 2D joint locations}

\begin{table}[tb]
	\begin{center}
		\scalebox{0.635}{
			\begin{tabular}{lcccccccccccccccc}
				\toprule
				Method & Direct. & Discuss & Eating & Greet & Phone & Photo & Pose & Purch. & Sitting & SitingD & Smoke & Wait & WalkD & Walk & WalkT & Avg \\
				\midrule
				\textbf{w/ 3D dataset} & & & & & & & & & & & & & & & &\\ 
				~~~Tang et al.~\cite{tung2017adversarial}          &  53.7 &  71.5 &  82.3 &  58.6 &  86.9 &  98.4 &  57.6 & 104.2 & 100.0 & 112.5 &  83.3 &  68.9 &   --- &  57.0 &   --- &  79.0 \\
				~~~Martinez et al.~\cite{martinez20173dbaseline}   &  37.7 &  44.4 &  40.3 &  42.1 &  48.2 &  54.9 &  44.4 &  42.1 &  54.6 &  58.0 &  45.1 &  46.4 &  47.6 &  36.4 &  40.4 &  45.5 \\ \midrule
				\textbf{w/o 3D dataset} & & & & & & & & & & & & & & & &\\ 
				~~~Ours                                            & 125.0 & 137.9 & 107.2 & 130.8 & 115.1 & 127.3 & 147.7 & 128.7 & 134.7 & 139.8 & 114.5 & 147.1 & 130.8 & 125.6 & 151.1 & 130.9 \\
				\bottomrule
			\end{tabular}
		}	
	\end{center}
	\caption{Average error in millimeters between the predicted 3D pose and the ground truth 3D pose using the ground truth 2D pose as input. While the methods except for ours use a 3D dataset during training, our method does not use a 3D dataset, and is performed in a completely unsupervised manner.}
	\label{gt2d}
\end{table}

When estimating a 3D pose from an image using our method, we require 2D joint locations from a 2D joint detector.
Thus, when evaluating our method as a 3D pose estimator from an image, the error naturally depends on the quality of the detected 2D joints.
The best score can be achieved when the ground truth 2D joint locations are used as inputs.
Therefore, we first evaluated our method using the ground truth 2D joint locations.
Some studies~\cite{martinez20173dbaseline,tung2017adversarial} that aim to lift 2D to 3D using 3D datasets during training had also used ground truth 2D joint locations for evaluation.

\Tref{gt2d} shows the results of 3D pose estimation from 2D joint locations.
The result of our unsupervised approach shows that the error is higher than the state-of-the-art method~\cite{martinez20173dbaseline} that uses 3D datasets and trains neural networks similarly to ours in a supervised manner.
Because the primary difference between our method and \cite{martinez20173dbaseline} is whether 3D annotations were used during training, the difference in these errors are primarily due to the 3D supervision.
The supervision of a 3D pose from 2D joint locations includes imaging parameters such as scale, distance from camera, and camera angle.
Unlike \cite{martinez20173dbaseline}, we assume that the cameras are placed horizontally because the camera angles are unknown in our method.
Thus, we speculate that the error in our method is primarily caused by the lack of camera angles from 3D supervision.

\subsection{Training on 2D detection}

\begin{table}[tb]
	\begin{center}
		\scalebox{0.635}{
			\begin{tabular}{lcccccccccccccccc}
				\toprule
				Method & Direct. & Discuss & Eating & Greet & Phone & Photo & Pose & Purch. & Sitting & SitingD & Smoke & Wait & WalkD & Walk & WalkT & Avg \\
				\midrule
				\textbf{w/ 3D dataset} & & & & & & & & & & & & & & & &\\ 
			~~~Ionescu et al.~\cite{ionescu2014human3.6m}            & 132.7 & 183.6 & 132.3 & 164.4 & 162.1 & 205.9 & 150.6 & 171.3 & 151.6 & 243.0 & 162.1 & 170.7 & 177.1 &  96.6 & 127.9 & 162.1 \\
			~~~Tung et al.~\cite{tung2017adversarial}        &  77.6 &  91.4 &  89.9 &  88.0 & 107.3 & 110.1 &  75.9 & 107.5 & 124.2 & 137.8 & 102.2 &  90.3 &  ---  &  78.6 &  ---  &  97.2 \\
			~~~Tome et al.~\cite{tome2017lifting}            &  65.0 &  73.5 &  76.8 &  86.4 &  86.3 & 110.7 &  68.9 &  74.8 & 110.2 & 173.9 &  85.0 &  85.8 &  86.3 &  71.4 &  73.1 &  88.4 \\
			~~~Pavlakos et al.~\cite{pavlakos2017coarse}     &  67.4 &  72.0 &  66.7 &  69.1 &  72.0 &  77.0 &  65.0 &  68.3 &  83.7 &  96.5 &  71.7 &  65.8 &  74.9 &  59.1 &  63.2 &  71.9 \\
			~~~Martinez et al.~\cite{martinez20173dbaseline} &  51.8 &  56.2 &  58.1 &  59.0 &  69.5 &  78.4 &  55.2 &  58.1 &  74.0 &  94.6 &  62.3 &  59.1 &  65.1 &  49.5 &  52.4 &  62.9 \\
			~~~Zhou et al.~\cite{zhou2017towards}            &  54.8 &  60.7 &  58.2 &  71.4 &  62.0 &  65.5 &  53.8 &  55.6 &  75.2 & 111.6 &  64.2 &  66.1 &  51.4 &  63.2 &  55.3 &  64.9 \\ \midrule
				\textbf{w/o 3D dataset} & & & & & & & & & & & & & & & &\\ 
				~~~Ours                                            & 161.3 & 174.3 & 143.1 & 169.2 & 161.7 & 174.1 & 180.7 & 178.0 & 170.6 & 191.4 & 157.4 & 182.3 & 180.7 & 180.3 & 193.4 & 173.2 \\
				\bottomrule
			\end{tabular}
		}	
	\end{center}
	\caption{Average error in millimeters between predicted 3D pose and ground truth 3D pose. Our method use detected 2D joint locations by Stacked Hourglass as inputs. All methods except for ours use a 3D dataset during training.}
	\label{h36m}
\end{table}

\begin{figure}[tb]
	\begin{center}
		\includegraphics[width=11cm]{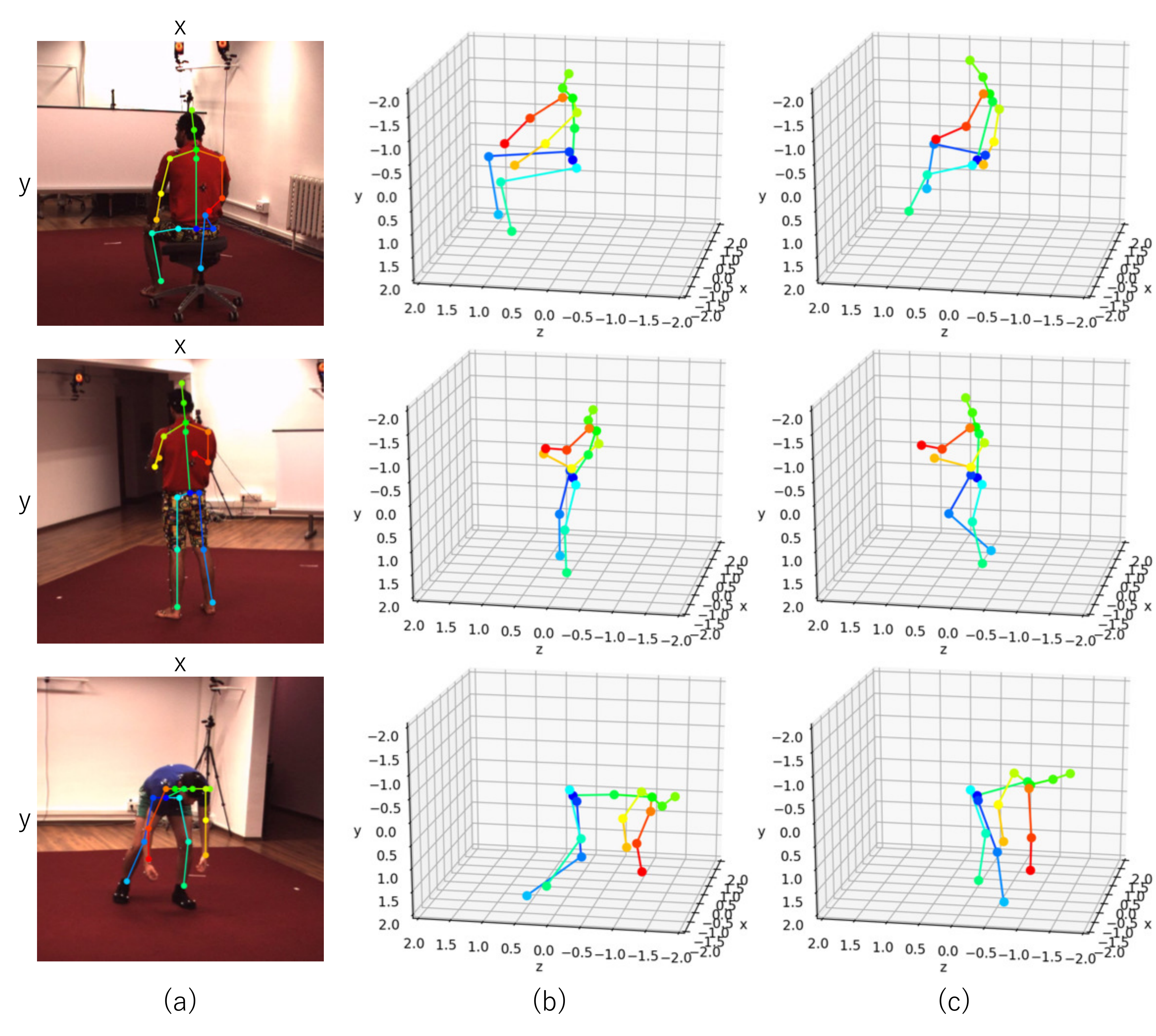}
		\caption{Qualitative results on the Human3.6M dataset using detected 2D joint locations as input. Column (a) denotes the input 2D joint locations, column (b) denotes the ground truth 3D pose in a relative space, and column (c) denotes the predicted 3D pose in a relative space.}
		\label{com}
	\end{center}
\end{figure}

\begin{figure}[tb]
	\begin{center}
		\includegraphics[width=12cm]{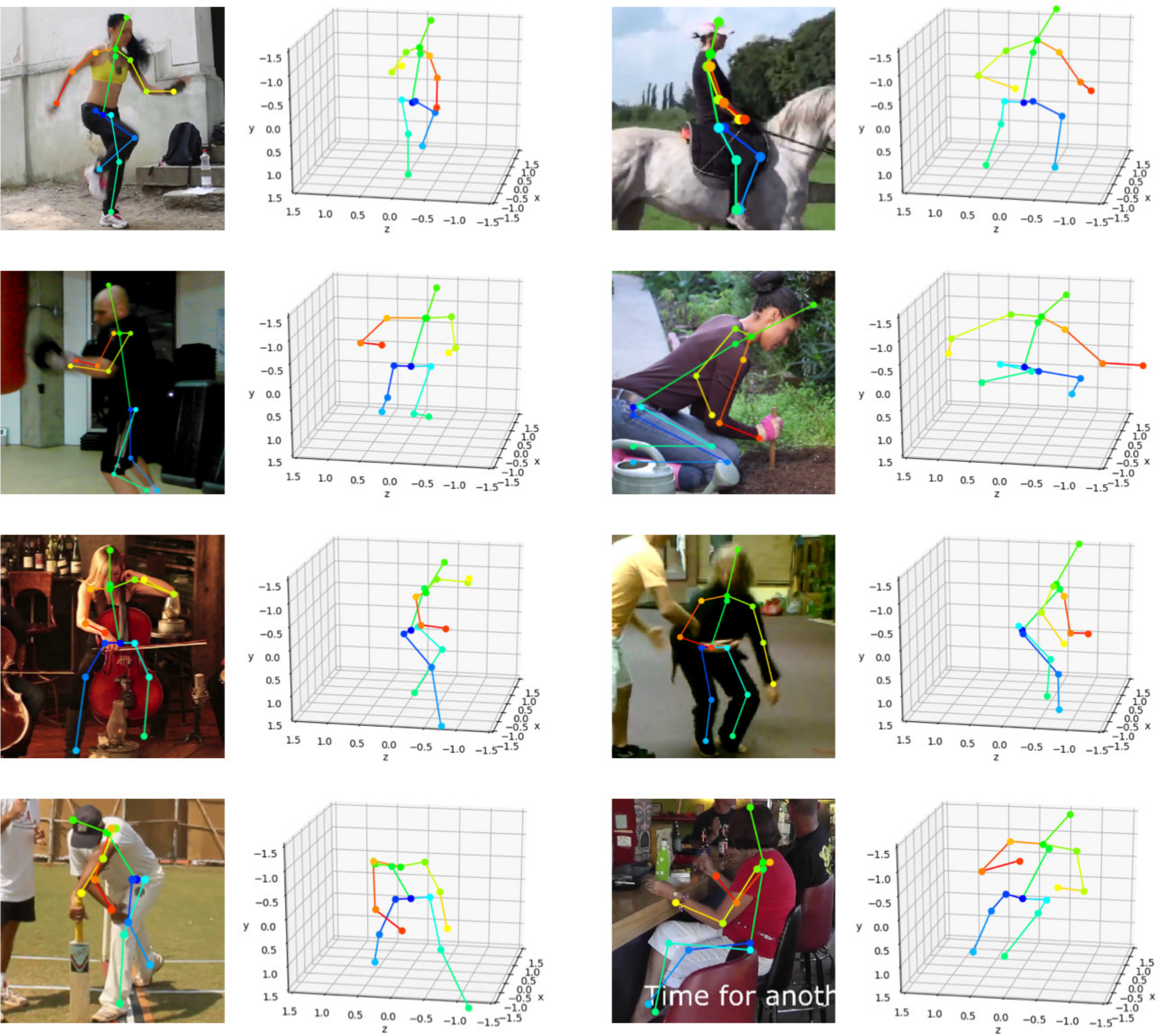}
		\caption{Qualitative results of the MPII dataset using detected 2D poses by Stacked Hourglass.}
		\label{mpii-result}
	\end{center}
\end{figure}

We evaluated the error of 3D pose estimation using the results from a 2D joint detector (not the ground truth 2D joints) as inputs.
We used the Stacked Hourglass~\cite{newell2016stacked} as a 2D joint detector and trained our networks using the detected 2D joint locations.
The Stacked Hourglass can detect 2D joints efficiently and accurately.
Some research works~\cite{pavlakos2017coarse,martinez20173dbaseline} about 3D human pose estimation also use the Stacked Hourglass as a 2D joint detector.
\Tref{h36m} shows the results of the average error in millimeters between the ground truth 3D pose and estimated 3D pose from images.
We found that the error in our method is higher than that of the existing approach that use a 3D dataset.
We speculate that the loss of error is also caused by the lack of imaging parameters that can be obtained from the 3D dataset mentioned in section 4.3.

\Fref{com} shows the qualitative results of our method using a test set from the Human3.6M dataset.
Our system can generate very natural 3D human poses even when a 3D dataset is not available during training.
Because the images in the Human3.6M dataset were captured by the cameras that were not placed horizontally, the predicted 3D poses were rotated around the $x$-axis compared with the ground 3D pose.

We also tested our method using the MPII dataset that has only 2D annotations.
Although the existing methods that use 3D datasets have estimated 3D locations of the joints provided by the 3D datasets, we estimated those that were provided by the MPII dataset.
\Fref{mpii-result} shows the qualitative results of the MPII.
We found that our method also performs well for a common 2D pose dataset, which contains images captured in the wild.
Therefore, our method can be applied to 3D pose estimation in the wild using a 2D joint detector.

\section{Discussion and Future Work}
\subsection{Analysis of error}
If we train a neural network to estimate a 3D pose from 2D joint locations in a supervised manner, we generally use a 2D and 3D pair, and this implies that some of the camera parameters, such as the translation, rotation, focal length, and distortion can be obtained while training.
In an unsupervised manner, the camera parameters are not obtained because a 2D and 3D pair is not available.
Moreover, it is unrealistic to expect these camera parameters explicitly during training.

In our approach, we assume that an input 2D pose is a projection of a specific 3D pose using orthographic cameras that are horizontally placed.
Based on these assumptions, given a 2D pose, we can generate the corresponding $z$-components for each joint in a 3D space relative to the central human joint.
However, most existing cameras are not orthographic cameras and are not placed horizontally.
Because the images in the Human3.6M dataset are also not captured by orthographic cameras horizontally, the 3D pose estimated by our method contains errors caused by these failed assumptions.

Because the cameras used in the Human3.6M dataset are not orthographic cameras, inevitable errors occurred between the captured 2D pose by the cameras in Human3.6M and the projected ground truth 3D pose using orthographic cameras.
Fortunately, the errors are not large because the variance of the depth of human joints is relatively small.
Generally, the difference between an image captured by a perspective camera and an image captured by an orthographic camera depends on the variance of the depth of the captured object.

Another factor that causes inevitable errors from the failed assumption is that the cameras in Human3.6M are not placed horizontally.
The errors increase as the horizontal angle of the cameras become bigger.
Therefore, our method does not work well if an image is captured from a strange angle such as a bird's eye view camera.
To avoid these problems, the method to estimate camera parameters from its images is very helpful for our work, and this would be the future work to improve our method.

\subsection{Applications to other tasks}
Our approach can be directly applied to a semi-supervised learning.
Let us suppose we have a dataset that contains many 2D joint locations, but does not contain 3D annotations, such as the MPII dataset.
If we annotate the 3D joint locations to some of the images in this dataset and train the neural network in a supervised manner, the unlabeled 2D joints can be helpful using our techniques.

Our techniques can also be applied to estimate the 3D structure of general objects when we cannot access the 3D data but can access large amounts of 2D data, such as the 2D images and 2D masks of the objects.

\section{Conclusion}
In this study, we proposed a simple yet effective method that predicts a 3D human pose from single 2D human joint locations.
Our system is based on the generative adversarial networks that are trained in an unsupervised manner, without any 3D human pose datasets.
Our primary idea was that, if the network can predict a 3D human pose correctly, the 3D pose projected onto a 2D plane should not collapse even if it is rotated perpendicularly.

We tested our method using the Human3.6M dataset and showed that our network can predict a 3D pose from single 2D joint locations accurately.
We also showed that our method can be easily applicable for the 3D human pose estimation from a single image by combining with a 2D joint detector.
We tested the approach that uses our method and a 2D joint detector using the MPII dataset, which has no 3D supervision and has many images in the wild, and showed that this approach is effective for such datasets.

\bibliographystyle{splncs}
\bibliography{egbib}


\title{Supplementary Material} 

\titlerunning{Unsupervised Adversarial Learning of 3D Human Pose}

\authorrunning{Kudo et al.}

\author{\empty}
\institute{\empty}

\maketitle

\section{Additional qualitative results}
\begin{figure}[tb]
  \begin{center}
    \includegraphics[width=12cm]{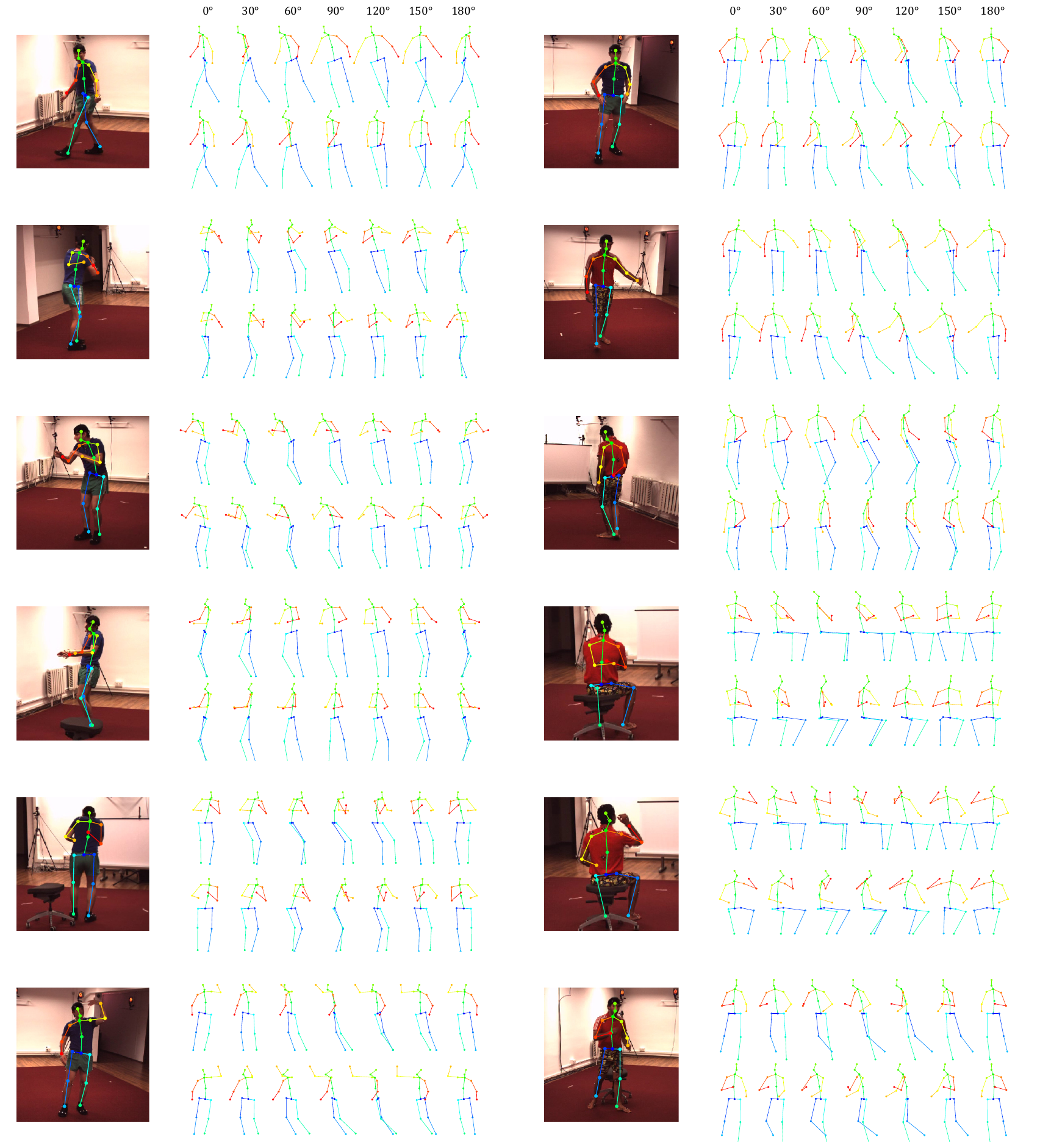}
    \caption{
    Results of the Human3.6M dataset~\cite{S_h36m}.
    The upper poses are the ground truth 3D poses.
    The lower poses are the predicted 3D poses from ground truth 2D joint locations.
    }
    \label{h36m_gt}
  \end{center}
\end{figure}

\begin{figure}[tb]
  \begin{center}
    \includegraphics[width=12cm]{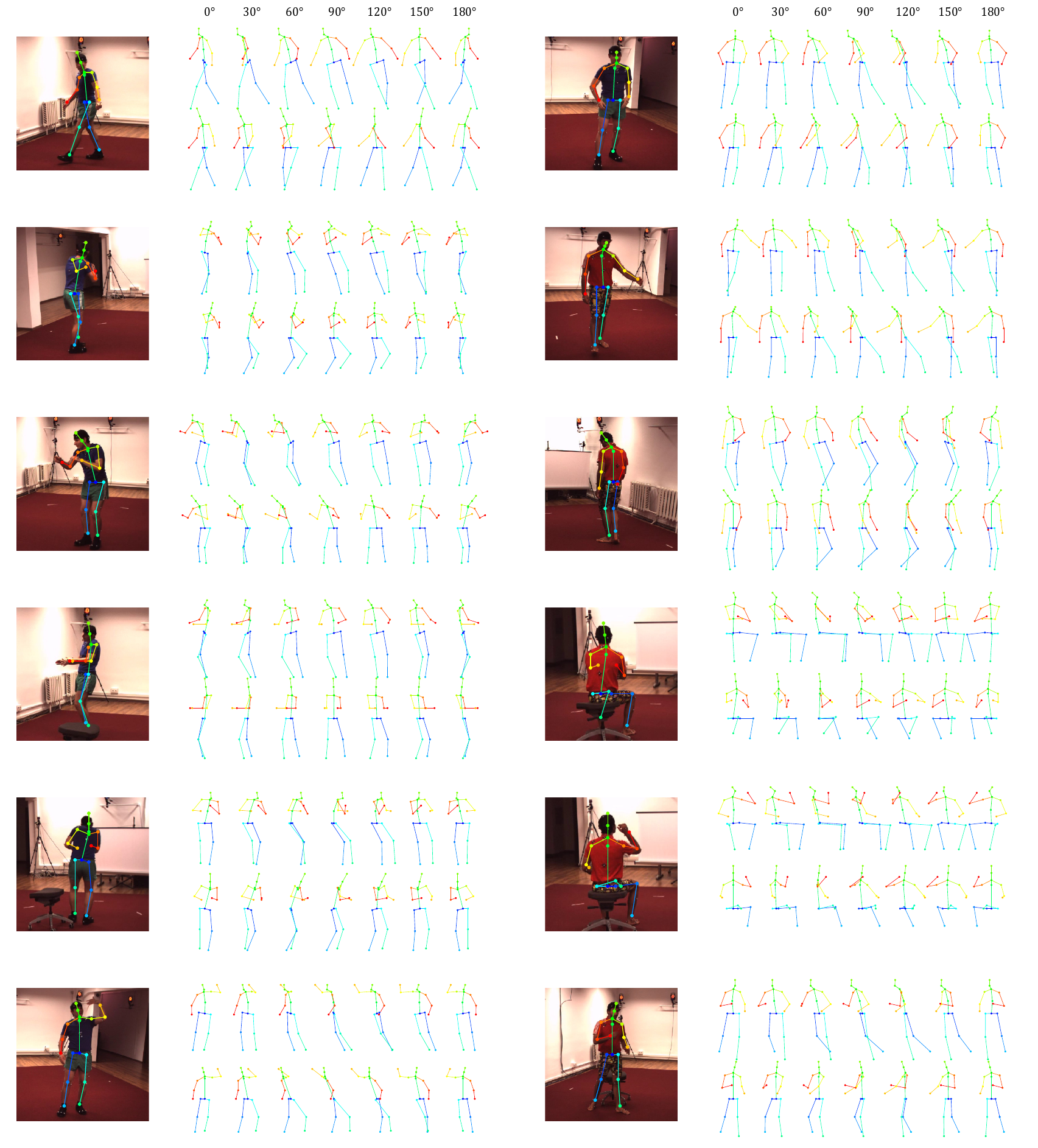}
    \caption{
    Results of the Human3.6M dataset~\cite{S_h36m}.
    The upper poses are the ground truth 3D poses.
    The lower poses are the predicted 3D poses from detected 2D joint locations by Stacked Hourglass~\cite{S_newell}.
    }
    \label{h36m_sh}
  \end{center}
\end{figure}

\begin{figure}[tb]
  \begin{center}
    \includegraphics[width=12cm]{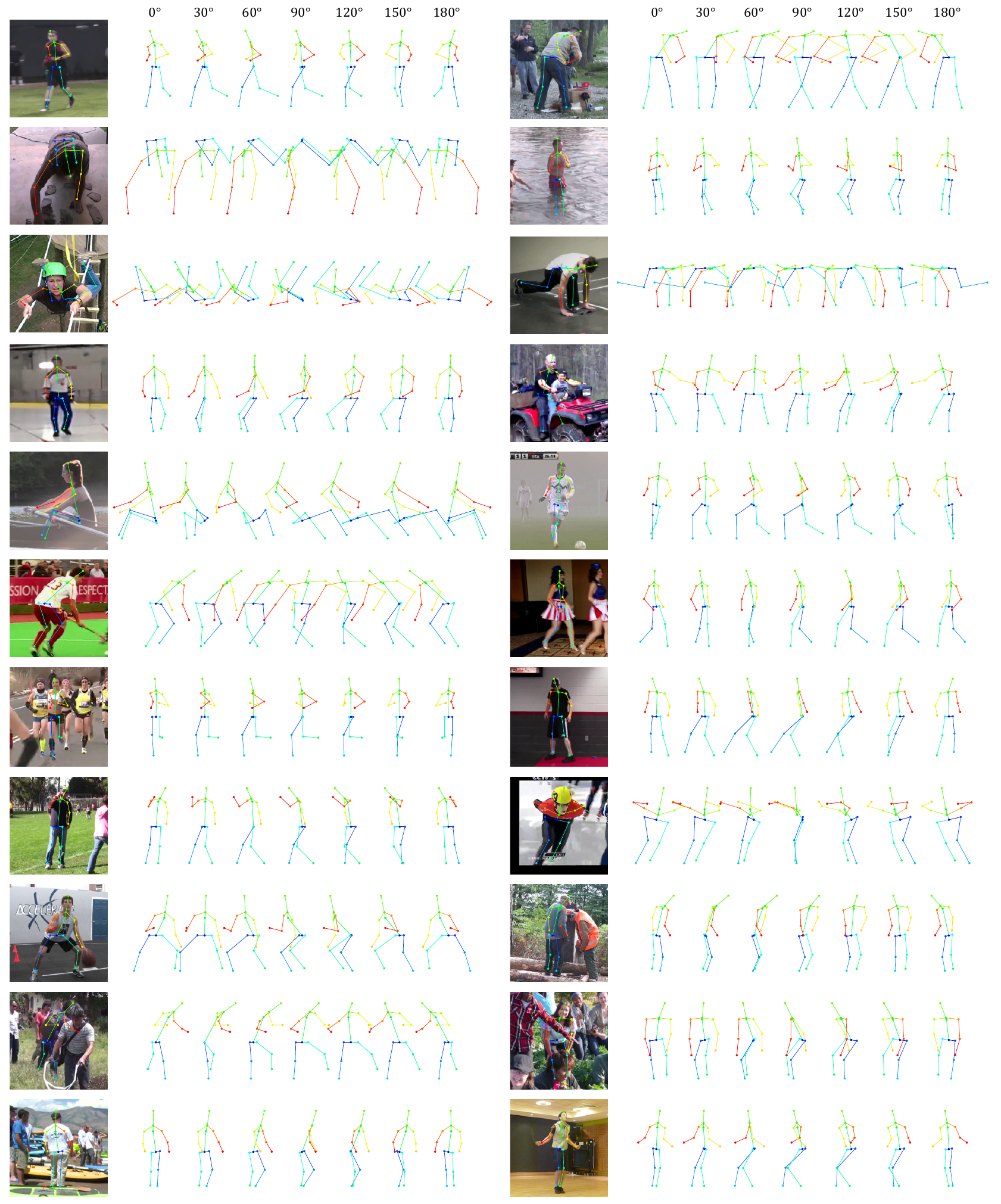}
    \caption{
    Results of the MPII dataset~\cite{S_mpii}.
   	The 3D poses are predicted from ground truth 2D joint locations.
    }
    \label{mpii_gt}
  \end{center}
\end{figure}

\begin{figure}[tb]
  \begin{center}
    \includegraphics[width=12cm]{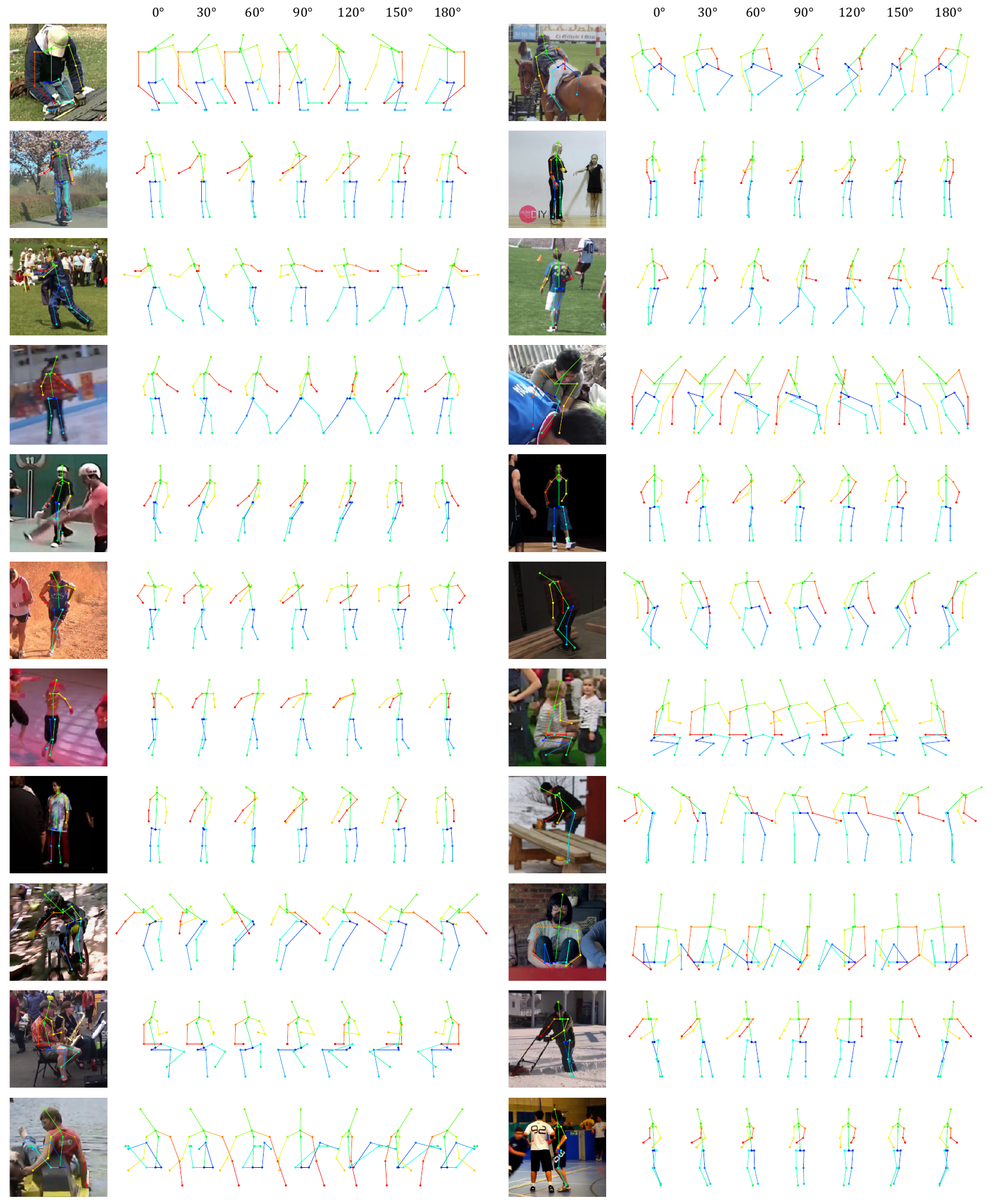}
    \caption{
    Results of the MPII dataset~\cite{S_mpii}.
   	The 3D poses are predicted from detected 2D joint locations by Stacked Hourglass~\cite{S_newell}.
    }
    \label{mpii_sh}
  \end{center}
\end{figure}

We show the additional results of our system.
\Fref{h36m_gt} and \Fref{mpii_gt} show the predicted 3D poses from the ground truth 2D joint locations.
\Fref{h36m_sh} and \Fref{mpii_sh} show the predicted 3D poses from the detected 2D joint locations by Stacked Hourglass~\cite{S_newell}.
The 3D poses in these figures are rotated around the vertical axis by every 30 degree in the clockwise direction.

The predicted 3D poses are not collapse from any view points around the vertical axis if the inputs are the ground truth 2D joint locations.
Furthermore, our system works well when we use detected 2D poses at train and test time.

Our method can learn to estimate 3D poses with an existing 2D pose dataset itself.
The predicted 3D poses for MPII dataset will be publicly available as a new 3D pose dataset.

\section{Evaluation for MPI-INF-3DHP dataset}
Meta et al.~\cite{S_mpi-inf} proposed a new 3D pose dataset, which is known as MPI-INF-3DHP.
The images in this dataset are captured by a mocap system in indoor and outdoor scenes.
Meta et al. trained convolutional neural networks to estimate 3D poses {\it from a single image} using this dataset.
They reported $84.1$ PCK with a threshold $150mm$ in a green screen background scene.
We trained our networks using 2D joint locations from the MPI-INF-3DHP dataset.
We evaluated our method that predict 3D human poses {\it from ground truth 2D joint locations} and report $89.3$ PCK under the same conditions.
This result implies that good 2D pose estimation enables us to predict 3D poses from an image accurately.

\end{document}